\documentclass{article}

\usepackage{microtype}
\usepackage{graphicx}
\usepackage{booktabs}
\usepackage{hyperref}

\usepackage[accepted]{icml2025}
\makeatletter
\renewcommand{\Notice@String}{Preprint. Copyright 2026 by the author(s).}
\makeatother

\usepackage{amsmath}
\usepackage{amssymb}
\usepackage{mathtools}
\usepackage{amsthm}

\usepackage{algorithm}
\usepackage{algorithmic}
\usepackage{multirow}
\usepackage{makecell}
\usepackage{subcaption}
\usepackage{float}
\usepackage{xcolor}
\usepackage{tikz}
\usetikzlibrary{arrows.meta,positioning,shapes.geometric}

\theoremstyle{plain}

\theoremstyle{definition}

\theoremstyle{remark}

\newcommand{\lucid}{\textsc{LUCID}}
\newcommand{\cdnots}{\textsc{CDNOTS}}
\newcommand{\RR}{\mathbb{R}}

\icmltitlerunning{\lucid{}: Learning Under Confounding in Time Series}

\begin{document}

\twocolumn[
\icmltitle{\lucid{}: Learning Under Confounding for Inference and
           Discovery in Time Series}

\begin{icmlauthorlist}
\icmlauthor{Mohammad Fesanghary}{bbg}
\end{icmlauthorlist}
\icmlaffiliation{bbg}{Bloomberg L.P}
\icmlcorrespondingauthor{Mohammad Fesanghary}{mfesanghary1@bloomberg.net}

\icmlkeywords{causal discovery, latent confounding, time series, spectral methods}
\vskip 0.3in
]

\printAffiliationsAndNotice{}

\begin{abstract}
Unobserved common causes are pervasive in real-world time series and can induce spurious
associations that causal discovery methods mistake for direct edges. We propose \lucid{}
(\textbf{L}earning \textbf{U}nder \textbf{C}onfounding for \textbf{I}nference and
\textbf{D}iscovery), a regime-adaptive deconfounding layer that first estimates the
confounding regime
from data using a Mar\v{c}enko--Pastur spectral router, then applies a deconfounding
strategy matched to that regime. When the spectrum indicates pervasive factor
confounding, \lucid{} attenuates factor-dominated variation and recovers contemporaneous
(lag-$0$) structure from the resulting innovations, with edge selection calibrated
against a data-driven edge-free null. Rather than being tied to a particular discovery
algorithm, it can wrap existing discovery
engines; we demonstrate consistent improvements across three such methods. On a diverse
synthetic out-of-distribution benchmark spanning changes in confounder strength and sparsity,
loading density, lag structure, volatility dynamics, edge heterogeneity, persistence,
intermittency, and tail behavior, \lucid{} achieves the best family-weighted
\emph{directed, lag-resolved} graph $F_1$ ($0.60$), improving over the strongest baseline
by $0.19$ absolute ($\approx\!46\%$ relative). Its advantage widens relative to looser
lag-collapsed scoring, and remains robust under intermittent and heavy-tailed
confounding. Code reproducing the method, the benchmark generators, and every
reported experiment is available at \url{https://github.com/bloomberg/causal-ts}.
\end{abstract}
\section{Introduction}
\label{sec:intro}

Causal discovery from multivariate time series is critical across climate science,
finance, neuroscience, and industrial monitoring~\citep{peters2017elements,
runge2019detecting, spirtes2000causation}. A pervasive obstacle is latent
confounding---when observed variables are driven by unrecorded common causes: unmeasured
regional rainfall might affect multiple river gauges, a macroeconomic shock might move
disparate assets, or a shared environment could regulate several genes. These hidden drivers
induce spurious associations that mimic direct causal effects, so discovery algorithms that
ignore them systematically over-connect the causal graph.

Latent confounding lacks a universal statistical signature, so no single correction
handles all of its forms. Under \emph{pervasive} confounding, a few latent factors affect
large subsets of the observed variables, which leaves two exploitable traces: a low-rank
component in the covariance, whose leading eigenvalues separate from the bulk, and
algebraic rank constraints among the many observed children of each factor. Spectral and
factor-adjustment methods~\citep{cevid2020spectral} use the former, while rank- and
tetrad-based methods~\citep{spirtes2000causation, dong2024rlcd} use the latter. Under
\emph{sparse} confounding, hidden causes are localized, for example a single latent fork
driving only two or three variables. Such a fork need not produce a separated eigenvalue,
and may provide too few observed children for informative rank constraints, so the
signals these corrections rely on can be weak or absent and genuine edges risk being
pruned. Corrections optimized for one regime often
degrade the other, so a robust algorithm must first detect \emph{which} regime is present.

\paragraph{Our approach.}
Rather than proposing a new discovery engine, \lucid{} is a
\emph{regime-adaptive deconfounding layer}: it infers from the observed data whether
confounding is sparse or pervasive, then routes each dataset to a correction appropriate
for that regime. The routing decision is based on a Mar\v{c}enko--Pastur spectral
statistic~\citep{marchenko1967distribution} whose threshold is derived from the data
dimension and length rather than tuned per dataset. On the pervasive branch, the
contemporaneous edges that a shared factor would otherwise smear into a dense clique are
\emph{recovered} by a spectral low-rank-plus-sparse step with a self-calibrating threshold.
Because the router and the recovery step both act on the data and on the returned
graph---never on the internals of the search---\lucid{} can wrap different discovery
engines rather than replacing them. We demonstrate this with
\cdnots{}~\citep{fesanghary2025cdnots} (our default),
PCMCI+~\citep{runge2020discovering}, and NTS-NOTEARS~\citep{sun2023nts} alike.

\paragraph{Contributions.}
\begin{enumerate}
\item A \emph{regime router} for latent confounding grounded in the
Mar\v{c}enko--Pastur null distribution (Section~\ref{sec:router}). Its threshold is
derived from $d$ and $T$ rather than tuned, and it recovers most of the value of an oracle
that picks the branch per dataset (regret $0.007$; Section~\ref{sec:diagnostics}).
Its role is robustness: committing to the always-sparse branch costs $0.26$
family-weighted $F_1$ relative to routing, while \lucid{} avoids that loss without
knowing the confounding regime in advance.

\item A \emph{spectral low-rank-plus-sparse (S$-$L) recovery} of the
contemporaneous (lag-$0$) slice on the pervasive branch
(Section~\ref{sec:sl}). The procedure attenuates factor-dominated variation and
reconstructs sparse contemporaneous candidates, calibrating its edge threshold against an
edge-free null built from the data rather than a hand-tuned amplitude constant. A
\emph{persistence gate} further suppresses the co-parent (moralization) edges that a
precision-matrix support would otherwise admit.

\item An evaluation suite of ten confounding families spanning the axes listed above,
together with a case study whose confounding dynamics are calibrated to a real
credit-default-swap panel (Section~\ref{sec:cds}). Scored throughout with the strict directed, lag-resolved
$F_1$, \lucid{} reaches $0.601$ against $0.411$ for the strongest baseline
(Table~\ref{tab:ood}); its margin \emph{grows} under lag-resolved rather than
lag-collapsed scoring, and the layer transfers: wrapped around PCMCI+ or NTS-NOTEARS
instead of \cdnots{}, it improves each of them.
\end{enumerate}

\section{Related Work}
\label{sec:related}

\paragraph{Time-series causal discovery.}
A broad class of methods learns temporal causal structure under versions of causal
sufficiency. Constraint-based approaches such as PC and its time-series
extensions~\citep{spirtes2000causation}, PCMCI/PCMCI+~\citep{runge2019detecting,
runge2020discovering}, and \cdnots{}~\citep{fesanghary2025cdnots} construct a candidate
graph and prune it using conditional-independence tests before orienting the remaining
edges. Score-based and continuous-optimization approaches instead search for a graph
that optimizes a global objective: DYNOTEARS~\citep{pamfil2020dynotears} and
NTS-NOTEARS~\citep{sun2023nts} extend NOTEARS-style optimization to temporal structure,
while VARLiNGAM~\citep{hyvarinen2010estimation} exploits non-Gaussianity to identify a
linear structural ordering. Neural approaches such as Neural Granger
Causality~\citep{tank2021neural}, NAVAR~\citep{bussmann2021neural}, and
CUTS+~\citep{cheng2023cuts_plus} infer predictive causal structure with nonlinear
function classes. These methods differ substantially in their search procedures and
assumptions, making them useful test beds for our central goal: a deconfounding layer
that can improve an existing discovery engine without modifying its internal search.

\paragraph{Discovery in the presence of latent confounders.}
A separate line of work relaxes causal sufficiency explicitly. FCI~\citep{spirtes2000causation}
and RFCI~\citep{colombo2012learning}, together with time-series extensions including
tsFCI~\citep{entner2010causal}, SVAR-FCI~\citep{malinsky2018causal},
LPCMCI~\citep{gerhardus2020high}, and TS-ICD~\citep{rohekar2023temporal}, reason about
latent common causes through partial ancestral graphs and edge marks that encode causal
ambiguity. These methods therefore
address latent confounding by changing the graphical object being recovered. Rank-based
approaches take a different route: methods such as RLCD~\citep{dong2024rlcd} exploit
rank constraints and vanishing tetrads to infer latent structure directly from observed
covariances. A third route recovers the latent components themselves and enlarges the
observed system with them: LCCI~\citep{aghaei2026lcci} applies time-delay embedding and
independent component analysis to extract latent causal components, retains those with
system-wide influence, and appends them to the observed series before an unmodified
downstream discovery algorithm is run. These paradigms aim to represent or reconstruct
hidden causal structure,
whereas \lucid{} targets a different objective: recovering a directed, lag-resolved graph
over the observed variables while treating latent confounding as a nuisance that must be
detected and corrected around the underlying discovery engine.

\paragraph{Factor, rank, and low-rank deconfounding.}
The closest methodological relatives of \lucid{} are methods that exploit the statistical
signature left by pervasive latent factors. When a small number of hidden variables affect
many observed variables, their contribution can appear as low-rank structure in the
covariance or precision matrix. Rank- and tetrad-based methods exploit the resulting
algebraic constraints~\citep{dong2024rlcd}, while factor-adjustment methods estimate latent
components and condition on them before downstream inference~\citep{wang2019blessings}.
Low-rank-plus-sparse formulations instead decompose observed dependence into a shared
low-rank component and a sparse graphical component~\citep{chandrasekaran2012latent,
frot2019robust}. Closest to the spectral step used here, \citet{cevid2020spectral}
deconfound by transforming the singular values of the design matrix before fitting a
Lasso, and DeCAMFounder~\citep{agrawal2023decamfounder} uses spectral decomposition for
causal discovery under pervasive latent variables; both target i.i.d.\ observations
rather than time series, and the former estimates regression coefficients rather than
graph structure. \lucid{} builds on this lineage: on the pervasive branch, its S$-$L
procedure recovers contemporaneous structure rather than discarding the lag-$0$ slice,
though it does so with a cheaper spectral procedure than the convex low-rank-plus-sparse
program it is inspired by (Section~\ref{sec:sl}). The limitation of applying such
corrections universally motivates our router. Their identifying signal is strongest when
latent confounding is sufficiently pervasive to create a detectable spectral or rank
signature; a sparse latent fork affecting only a few observed variables need not do so,
and factor-style correction can then remove genuine dependence.

\paragraph{Positioning \lucid{}.}
\lucid{} builds on the low-rank view of pervasive latent confounding, but changes both
\emph{when} and \emph{how} that structure is used. Existing factor-based corrections
typically commit to a factor-confounding model before downstream inference. \lucid{}
first tests whether the observed spectrum supports that assumption and invokes
factor-style correction only in the pervasive regime. Within that branch,
S$-$L attenuates rather than removes the dominant factor directions, then reconstructs
lag-$0$ candidates using a data-calibrated edge-free null and marginal corroboration
(Section~\ref{sec:sl}). This preserves potentially genuine contemporaneous signal while
suppressing both factor footprint and conditioning-induced co-parent associations. The
result is a separation between \emph{deconfounding} and \emph{causal search}: \lucid{}
adapts the former to the observed regime while leaving the underlying discovery engine
interchangeable.

\section{Background and Notation}
\label{sec:background}

We observe a stationary multivariate time series
$\mathbf{x}_t=(x_t^1,\dots,x_t^d)\in\RR^d$, $t=1,\dots,T$, and seek the lagged causal
graph $G\in\{0,1\}^{d\times d\times(L+1)}$, where $G[c,e,\ell]=1$ denotes a direct
effect $x^c_{t-\ell}\!\to\!x^e_t$ at lag $\ell\in\{0,\dots,L\}$. Our primary metric is
the \emph{directed, lag-resolved} $F_1$-score ($F_1^{\text{dir}}$), under which an edge
is correct only if its cause, effect, and lag all match the ground truth. We exclude
same-variable autoregressive edges from scoring because they are recovered reliably by
all methods in our experiments and would obscure errors in the cross-variable structure.
For completeness, Appendix~\ref{app:pairf1} reports the pair-level
$F_1^{\text{pair}}$, which collapses across lags and therefore credits the correct
variable pair even when its delay is wrong.

\paragraph{Latent confounding.}
Under pervasive confounding, a small number of latent factors affect many observed
variables~\citep{bernanke2005favar, stock2002forecasting}:
\[
x_t^i = \sum_k \lambda_{ik} f_t^k + (\text{sparse causal effects}) + \varepsilon_t^i.
\]
This induces a detectable low-rank component in the dependence structure. Sparse
(local-fork) confounding, where a latent cause affects only a few variables, need not
produce such a spectral signature. \lucid{} routes between these regimes using this
distinction (Section~\ref{sec:method}).

\section{Method}
\label{sec:method}

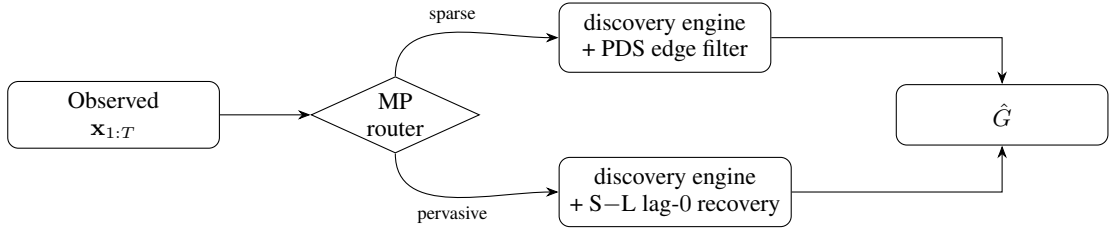
\begin{figure*}[t]
\centering
\begin{tikzpicture}[
  font=\small, >=Stealth, node distance=7mm and 12mm,
  box/.style={draw, rounded corners, align=center, minimum height=8mm, minimum width=28mm, inner sep=4pt},
  dec/.style={draw, diamond, aspect=2.2, align=center, inner sep=1pt},
  lbl/.style={font=\scriptsize}]
\node[box] (x) {Observed\\$\mathbf{x}_{1:T}$};
\node[dec, right=of x] (router) {MP\\router};
\node[box, above right=3mm and 16mm of router] (sparse) {discovery engine\\+ PDS edge filter};
\node[box, below right=3mm and 16mm of router] (perv) {discovery engine\\+ S$-$L lag-$0$ recovery};
\node[box, right=55mm of router] (out) {$\hat{G}$};
\draw[->] (x)--(router);
\draw[->] (router.north) to[out=90,in=180] node[lbl,above]{sparse} (sparse.west);
\draw[->] (router.south) to[out=-90,in=180] node[lbl,below]{pervasive} (perv.west);
\draw[->] (sparse.east) -| (out.north);
\draw[->] (perv.east) -| (out.south);
\end{tikzpicture}
\caption{\lucid{} pipeline. An MP spectral router selects a sparse or pervasive
deconfounding branch. Both use the same base discovery engine; the sparse branch applies
the PDS edge filter to every candidate edge, while the pervasive branch recovers the lag-$0$ slice with
S$-$L. We use \cdnots{} by default and also wrap PCMCI+ and NTS-NOTEARS.}
\label{fig:pipeline}
\end{figure*}

\lucid{} is a three-stage layer (Figure~\ref{fig:pipeline}): (i) infer the confounding
regime from the observed data; (ii) run the same base discovery engine on either branch
and apply the corresponding post-hoc correction; and (iii) on the pervasive branch,
replace the contemporaneous slice with S$-$L recovery. All hyperparameters are fixed
across datasets and experiments; Section~\ref{sec:params} gives their specification and
selection.

\subsection{Regime routing via the Mar\v{c}enko--Pastur null}
\label{sec:router}

We first fit a VAR(1) and form the correlation matrix of its residuals. Let
$\lambda_1\ge\lambda_2\ge\cdots\ge\lambda_d$ denote its eigenvalues and define
\[
R=\frac{\lambda_1+\lambda_2}{d},
\]
the fraction of residual variance captured by the top two spectral directions. Pervasive
factor confounding concentrates variance in a few directions and increases $R$, whereas
sparse local confounding need not produce such a spectral spike.

Rather than use a dataset-independent cutoff, we calibrate $R$ against the
Mar\v{c}enko--Pastur no-factor benchmark. With $T_{\text{eff}}=T-1$ residual samples, the
upper spectral edge is
$(1+\sqrt{d/T_{\text{eff}}})^2$~\citep{marchenko1967distribution,
johnstone2001distribution}. We therefore route using
\begin{equation}
\tau = \gamma\,\frac{k\,\bigl(1+\sqrt{d/T_{\text{eff}}}\bigr)^2}{d},
\label{eq:mp}
\end{equation}
with $k=2$ to match the two eigenvalues entering $R$ and a fixed margin
$\gamma=1.3$ (Table~\ref{tab:params}). If $R\le\tau$, \lucid{} selects the sparse branch;
otherwise it selects the pervasive branch. Thus the cutoff adapts automatically to $d$
and $T$, while $k$ and $\gamma$ remain fixed across datasets. Under sufficient
conditions the spectral statistic separates the two regimes asymptotically; a full
statement and proof are given in the extended version.

\subsection{Regime-specific discovery and filters}
\label{sec:filters}

Our default base engine is \cdnots{}~\citep{fesanghary2025cdnots} with a
partial-correlation CI test. Importantly, \lucid{} runs the same single-phase
PC-style skeleton search on both branches; the regime affects only the subsequent
deconfounding step. This isolates the contribution of routing and correction from changes
to the underlying causal search.

On the sparse branch, all candidate edges---contemporaneous ones included---are filtered
using a post-double-selection (PDS) edge filter~\citep{belloni2014inference}: a
Granger-style test at lags $\ell\ge1$, and the corresponding contemporaneous partial
regression at $\ell=0$ (Appendix~\ref{app:pds}). On the pervasive branch, lagged edges are retained
from the base engine and the lag-$0$ slice is instead reconstructed by S$-$L
(Section~\ref{sec:sl}). Additional pervasive-branch lag filters and tetrad-based
alternatives are evaluated as baselines and ablations in Section~\ref{sec:experiments}.

\paragraph{Novel vs.\ adopted components.}
The Mar\v{c}enko--Pastur law and the low-rank-plus-sparse view of latent confounding are
established tools; our contribution is their use in a regime-adaptive causal discovery
layer. Specifically, new in \lucid{} are the MP-based routing rule and, within S$-$L
(Section~\ref{sec:sl}), the edge-free-null calibration and the persistence gate. The
lag-$0$ orientation step is a heuristic we adopt for the directed evaluation rather than
a contribution of this work (Appendix~\ref{app:orient}). The low-rank-plus-sparse
formulation~\citep{chandrasekaran2012latent} that S$-$L is inspired by, the spectral
trimming operator itself~\citep{cevid2020spectral}, the base engine
(\cdnots{}~\citep{fesanghary2025cdnots}), and the PDS edge
filter~\citep{belloni2014inference} are adopted from prior work.

\subsection{Contemporaneous recovery via spectral deconfounding}
\label{sec:sl}

Pervasive same-period factors are especially damaging at lag $0$: they induce dense
cross-sectional dependence among the innovations that can be mistaken for instantaneous
edges. Dropping the entire slice removes this footprint but also every genuine
contemporaneous edge, which costs substantial accuracy where such edges exist
(Appendix~\ref{app:lag0}). \lucid{} instead attenuates the dominant factor directions and
recovers sparse conditional-dependence candidates from the residual innovations. This
S$-$L step is inspired by latent-variable Gaussian graphical model
selection~\citep{chandrasekaran2012latent}, where marginalizing latent variables induces
sparse and low-rank structure in the observed concentration matrix. We do not solve the
corresponding convex program or inherit its guarantees; S$-$L is a cheaper spectral
procedure tailored to lag-$0$ recovery.

Let $\mathbf{U}$ be the centered VAR residuals with SVD
$\mathbf{U}=\sum_i \sigma_i u_i v_i^\top$. We estimate the factor count $\hat{k}$ using
the Mar\v{c}enko--Pastur rule in Appendix~\ref{app:factorcount} and cap the leading
singular values at $\sigma_{\hat{k}+1}$:
\[
\tilde{\sigma}_i=\min(\sigma_i,\sigma_{\hat{k}+1}),\qquad
\tilde{\mathbf U}=\sum_i \tilde{\sigma}_i u_i v_i^\top .
\]
This is the Trim transform of \citet{cevid2020spectral}, applied here with a different
cap: they trim at a fixed quantile (the median singular value) to fit a Lasso under dense
confounding in the i.i.d.\ setting, whereas we reuse the factor count $\hat{k}$ already
estimated for routing and feed the trimmed residuals to contemporaneous structure
recovery rather than to a regression. Trimming suppresses factor-dominated directions
without discarding them entirely. From
$\tilde{\mathbf S}=\operatorname{Cov}(\tilde{\mathbf U})$ we compute
\[
\hat\Theta=(\tilde{\mathbf S}+\rho I)^{-1},\qquad
\rho=10^{-3}\operatorname{tr}(\tilde{\mathbf S})/d,
\]
and the corresponding partial correlations
$\hat\rho_{ij}=-\hat\Theta_{ij}/\sqrt{\hat\Theta_{ii}\hat\Theta_{jj}}$.

\paragraph{Persistence gate.}
Partial correlation alone does not identify the contemporaneous skeleton. In a linear
structural model, conditioning can induce nonzero precision entries between co-parents of
a collider even when no direct edge connects them. We therefore require marginal
corroboration in addition to conditional association. Let
$m_{ij}=|\operatorname{Spearman}(\tilde u^i,\tilde u^j)|$ and define
\begin{equation}
s_{ij}
=
|\hat\rho_{ij}|
\left(1-e^{-(m_{ij}/\tau_g)^2}\right),
\qquad \tau_g=0.15 .
\label{eq:slscore}
\end{equation}
The multiplier approaches one for pairs with strong marginal support and suppresses
conditional associations with little marginal evidence. Spearman correlation reduces
sensitivity to heavy tails, and the same score is used for both the observed data and
the null calibration below. This gate is deliberately heuristic: co-parents that are
also marginally associated can survive, so S$-$L produces a lag-$0$ skeleton candidate
rather than an identification guarantee (Section~\ref{sec:limitations}).

\paragraph{Edge-free null calibration.}
Rather than choose an amplitude cutoff for $s_{ij}$, we calibrate it against a null
constructed from the data. Write
$\mathbf{U}_{\hat{k}}=\sum_{i\le\hat{k}}\sigma_i u_i v_i^\top$ for the estimated factor
component and $\mathbf{E}=\mathbf{U}-\mathbf{U}_{\hat{k}}$ for the remainder. For each of
$B$ null draws, we independently circularly shift every column of $\mathbf{E}$ and
recombine it with the unchanged factor component. This destroys contemporaneous
alignment among the idiosyncratic coordinates while preserving the factor component and
each coordinate's serial structure.

For each draw we rerun S$-$L and compute the maximum null score
$\max_{i<j}s_{ij}$. The $(1-\alpha)$ quantile of these maxima defines
$\eta_\alpha$, and we retain $(i,j)$ only if $s_{ij}>\eta_\alpha$. Thus the threshold is
a data-driven max statistic rather than a tuned edge-amplitude constant. Under the
complete edge-free null and the shift-exchangeability assumption, the corresponding
randomization null targets family-wise error level $\alpha$; we use $\alpha=0.05$ and
$B=200$. This is a complete-null statement only: it does not exclude moralization errors
once genuine lag-$0$ edges are present. Strong conditional heteroskedasticity can also
violate shift exchangeability; for that setting we provide a more conservative stationary
block-bootstrap null~\citep{politis1994stationary}.

\paragraph{Orientation.}
Recovered lag-$0$ adjacencies are oriented using lead-lag asymmetry in the observed
series. For a candidate pair $(a,b)$, define
\[
A(a,b)=
\frac{
|\operatorname{corr}(a_t,b_{t+1})|
-
|\operatorname{corr}(b_t,a_{t+1})|
}{
|\operatorname{corr}(a_t,b_{t+1})|
+
|\operatorname{corr}(b_t,a_{t+1})|
}.
\]
We orient $a\!\to\!b$ when $A(a,b)>0$ and reverse the direction when $A(a,b)<0$.
The heuristic exploits temporal persistence following an instantaneous effect and, unlike
pairwise LiNGAM~\citep{hyvarinen2010estimation}, does not require non-Gaussian residuals.
Appendix~\ref{app:orient} states how the rule was selected and what it does not
establish.

\subsection{Fixed hyperparameters}
\label{sec:params}

Table~\ref{tab:params} lists all \lucid{} hyperparameters. They are fixed across datasets,
benchmarks, and the CDS case study; no dataset-specific tuning is performed. The S$-$L
edge threshold is calibrated from the edge-free null above and therefore introduces no
edge-amplitude parameter. The only empirically selected amplitude constant is the
persistence width $\tau_g$, chosen on lag-$0$ generators excluded from the evaluation
suite. Appendix~\ref{app:sensitivity} sweeps the PDS level, S$-$L null level, and
$\tau_g$, while Table~\ref{tab:router_sweeps} varies the router margin $\gamma$; the
reported results are stable across the tested ranges and the defaults are not selected
to maximize evaluation $F_1$.

\section{Experiments}
\label{sec:experiments}

\paragraph{Setup.}
We evaluate on synthetic generators with known ground truth and a market-calibrated
semi-synthetic CDS benchmark (Section~\ref{sec:cds}). Our constraint-based comparisons
isolate the vanishing-tetrad deconfounding filter as a design choice on top of the
single-phase, PC-style \cdnots{} skeleton engine. Table~\ref{tab:ood} includes
\cdnots{} and \cdnots{}$+$tetrad. We additionally compare with
VARLiNGAM~\citep{hyvarinen2010estimation}, DYNOTEARS~\citep{pamfil2020dynotears},
NTS-NOTEARS~\citep{sun2023nts}, PCMCI+~\citep{runge2020discovering}, and the
latent-confounder-aware LPCMCI~\citep{gerhardus2020high} and
TS-ICD~\citep{rohekar2023temporal}. All methods observe the same
data and are scored using the directed, lag-resolved $F_1^{\text{dir}}$. Unless noted,
synthetic results use $50$ seeds and a partial-correlation CI test where applicable.
Pair-level, lag-collapsed results are reported separately in
Appendix~\ref{app:pairf1}.

\subsection{Out-of-distribution generalization}
\label{sec:ood}

We evaluate across ten confounding families, weighted equally to reward robustness rather
than specialization to a single generator. Three families vary the topology of sparse
local confounding (Erd\H{o}s--R\'enyi, scale-free, and small-world), while seven pervasive
families vary volatility dynamics, structural effects, GARCH behavior, loading density,
lag structure, edge-weight heterogeneity, and factor persistence
(Appendix~\ref{app:families}). The dense-loading family is a deliberate stress test:
strong factor footprint coexists with a substantial number of genuine contemporaneous
edges, making indiscriminate lag-$0$ removal particularly costly.

Table~\ref{tab:ood} gives the main result. The default \lucid{} layer wrapped around
\cdnots{} reaches family-weighted
$F_1^{\text{dir}}=0.601{\pm}0.025$, compared with $0.411$ for the strongest unwrapped
baselines, an absolute gain of $0.19$ ($\approx\!46\%$ relative). The improvement is not
specific to \cdnots{}: wrapping PCMCI+ yields $0.596{\pm}0.025$, while wrapping
NTS-NOTEARS yields $0.538{\pm}0.023$. More importantly, each wrapped engine improves on
its own unwrapped counterpart in Overall score and in $27$ of the $30$
engine$\times$family comparisons. The three exceptions are PCMCI+ on pervasive-mixedlag
($0.273\!\to\!0.266$) and NTS-NOTEARS on Sparse-ER
($0.840\!\to\!0.804$) and Sparse-SW ($0.834\!\to\!0.829$).

The advantage also strengthens under the stricter lag-resolved metric. Relative to the
lag-collapsed pair-level score, \lucid{} drops only $0.018$ (from $0.619$), whereas the
baselines lose $0.04$--$0.08$, including $0.082$ for LPCMCI and $0.075$ for PCMCI+
(Appendix~\ref{app:pairf1}). Thus the gain is not merely in identifying the correct
variable pair: \lucid{} more often places the edge at the correct lag.

\begin{figure*}[t]
\centering
\includegraphics[width=\textwidth]{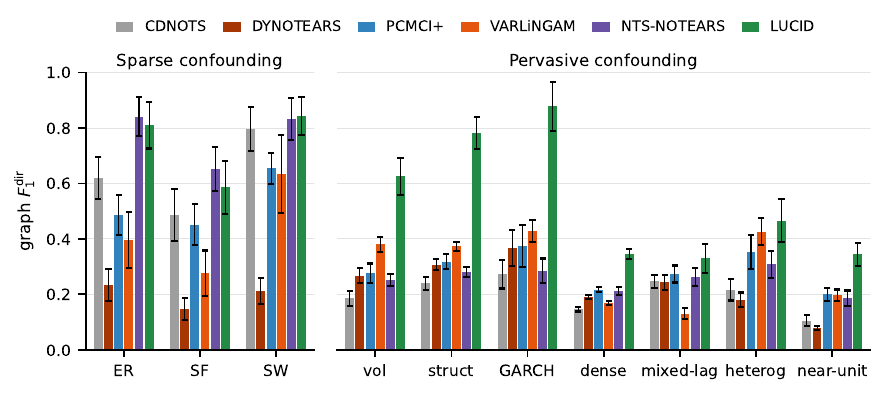}
\caption{Per-family out-of-distribution directed, lag-resolved graph-$F_1^{\text{dir}}$
($50$ seeds, mean$\pm$std), split by confounding regime. \lucid{} is the only method strong
in both: the baselines competitive on the sparse topologies fall to $\le\!0.43$ on every
pervasive family, while \lucid{} leads or ties the best of them on all seven, including the dense-loading
stress case. Values are those of Table~\ref{tab:ood}.}
\label{fig:byfamily}
\end{figure*}

The improvement is also not explained by the choice of base skeleton or by family
weighting. Applying the tetrad correction to the same single-phase skeleton leaves
Overall at $0.411$, while the default \lucid{} reaches $0.601$; throughout, the baseline
set excludes variants built on enhanced versions of our own base engine. Its sparse- and pervasive-regime macro-averages are
$0.746$ and $0.538$, respectively, giving $0.642$ under equal weighting of the two
regimes, against $0.515$ for the strongest unwrapped baseline (NTS-NOTEARS) under the same
weighting. The method is not uniformly best: NTS-NOTEARS remains stronger on some sparse
topologies, and dense, mixed-lag, and near-unit-root pervasive confounding remain difficult
for every method. The largest gains occur where pervasive confounding is most damaging:
on the structural family and the GARCH reference configuration, \lucid{} reaches $0.781$
and $0.878$, versus
$0.372$ and $0.429$ for the strongest baselines. Figure~\ref{fig:byfamily} gives the
full per-family profile.

\begin{table*}[t]
\centering
\caption{Out-of-distribution directed, lag-resolved graph
$F_1^{\text{dir}}$, family-weighted over ten confounding families and $50$ seeds.
Overall reports mean$\pm$std. The three sparse topologies are treated as separate
families rather than collapsed into one. Each \lucid{} row is placed beneath its
corresponding base engine. \textbf{Bold} marks the best method in each column and methods
not significantly worse than it (one-sided paired Wilcoxon signed-rank test,
$\alpha=0.05$). Pair-level results are in Appendix~\ref{app:pairf1}.}
\label{tab:ood}
\resizebox{\textwidth}{!}{%
\begin{tabular}{lccccccccccc}
\toprule
Method & Overall & Sparse-ER & Sparse-SF & Sparse-SW & Perv-vol & Perv-struct & GARCH & Perv-dense & Perv-mixedlag & Perv-heterog & Perv-nearunit \\
\midrule
\cdnots{}                & $0.331${\scriptsize$\pm.019$} & $0.619$ & $0.486$ & $0.796$ & $0.186$ & $0.239$ & $0.273$ & $0.146$ & $0.247$ & $0.217$ & $0.104$ \\
\cdnots{} + tetrad       & $0.411${\scriptsize$\pm.021$} & $0.621$ & $0.541$ & $0.797$ & $0.355$ & $0.352$ & $0.340$ & $0.148$ & $\mathbf{0.329}$ & $0.375$ & $0.254$ \\
\lucid{} (\cdnots{}, default) & $\mathbf{0.601}${\scriptsize$\pm.025$} & $0.810$ & $0.586$ & $0.843$ & $0.626$ & $\mathbf{0.781}$ & $\mathbf{0.878}$ & $0.345$ & $\mathbf{0.330}$ & $\mathbf{0.465}$ & $0.344$ \\
\midrule
VARLiNGAM                & $0.341${\scriptsize$\pm.020$} & $0.397$ & $0.276$ & $0.635$ & $0.381$ & $0.372$ & $0.429$ & $0.168$ & $0.131$ & $0.426$ & $0.197$ \\
\midrule
LPCMCI$^{\dagger}$       & $0.355${\scriptsize$\pm.023$} & $0.546$ & $0.403$ & $0.707$ & $0.287$ & $0.288$ & $0.301$ & --- & $0.274$ & $0.247$ & $0.146$ \\
TS-ICD$^{\ddagger}$      & $0.133${\scriptsize$\pm.024$} & $0.046$ & $0.007$ & $0.049$ & $0.352$ & $0.240$ & $0.317$ & $0.005$ & $0.045$ & $0.143$ & $0.121$ \\
PCMCI+                   & $0.360${\scriptsize$\pm.017$} & $0.486$ & $0.451$ & $0.655$ & $0.275$ & $0.317$ & $0.374$ & $0.216$ & $0.273$ & $0.353$ & $0.199$ \\
\lucid{} (PCMCI+)        & $\mathbf{0.596}${\scriptsize$\pm.025$} & $0.820$ & $0.556$ & $\mathbf{0.861}$ & $\mathbf{0.649}$ & $0.767$ & $0.829$ & $\mathbf{0.391}$ & $0.266$ & $\mathbf{0.472}$ & $\mathbf{0.354}$ \\
\midrule
DYNOTEARS                & $0.222${\scriptsize$\pm.010$} & $0.233$ & $0.147$ & $0.211$ & $0.267$ & $0.306$ & $0.367$ & $0.189$ & $0.244$ & $0.181$ & $0.079$ \\
\midrule
NTS-NOTEARS              & $0.411${\scriptsize$\pm.019$} & $\mathbf{0.840}$ & $0.653$ & $0.834$ & $0.251$ & $0.280$ & $0.284$ & $0.212$ & $0.263$ & $0.308$ & $0.186$ \\
\lucid{} (NTS-NOTEARS)   & $0.538${\scriptsize$\pm.023$} & $0.804$ & $\mathbf{0.690}$ & $0.829$ & $0.497$ & $0.579$ & $0.538$ & $0.318$ & $\mathbf{0.332}$ & $0.424$ & $\mathbf{0.371}$ \\
\bottomrule
\end{tabular}}

\vspace{2pt}
{\footnotesize $^{\dagger}$LPCMCI is not run on pervasive-dense ($d=24$; see
Section~\ref{sec:lpcmci}); its Overall score is family-weighted over the remaining nine
families. $^{\ddagger}$TS-ICD's PAG is projected with the same rule as LPCMCI's
(Section~\ref{sec:lpcmci}).}
\end{table*}

Table~\ref{tab:decomposition} shows that the gain is primarily precision-driven. The
three \lucid{} variants attain precision $0.491$--$0.614$ at recall $0.571$--$0.615$,
while the non-PAG baselines have precision $0.130$--$0.287$ at slightly higher recall
($0.596$--$0.721$): \lucid{} trades a little recall for a large gain in precision. The
same pattern appears in structural Hamming distance, where \lucid{} reaches
$13.4$--$14.5$ against $27.1$--$67.1$ for those baselines. The two PAG-based methods are
exceptions, for opposite reasons. LPCMCI reaches SHD $22.6$ at high recall ($0.671$) but
low precision ($0.233$), whereas TS-ICD attains SHD $14.0$---within the \lucid{}
range---at recall $0.123$, reflecting a far sparser recovered graph. SHD alone therefore
rewards a conservative output, whereas directed $F_1$ registers the structure such an
output fails to recover. This is consistent with the failure mode expected under
unobserved common causes: methods can retain substantial recall while adding many
confounder-induced false edges.

\begin{table}[t]
\centering
\caption{Precision, recall, and SHD underlying the OOD results, pooled over all datasets
and $50$ seeds. \lucid{}'s gain is primarily precision-driven, at slightly lower recall
than the baselines. Since $F_1$ is computed per dataset and then averaged, these pooled
precision and recall values do not reproduce Table~\ref{tab:ood} algebraically.}
\label{tab:decomposition}
\setlength{\tabcolsep}{4pt}
\begin{tabular}{lccc}
\toprule
Method & Precision & Recall & SHD \\
\midrule
\lucid{} (\cdnots{})   & $0.614$ & $0.571$ & $13.4$ \\
\lucid{} (PCMCI+)      & $0.593$ & $0.615$ & $14.5$ \\
\lucid{} (NTS-NOTEARS) & $0.491$ & $0.601$ & $13.6$ \\
\midrule
LPCMCI$^{\dagger}$   & $0.233$ & $0.671$ & $22.6$ \\
TS-ICD               & $0.168$ & $0.123$ & $14.0$ \\
PCMCI+               & $0.220$ & $0.641$ & $31.2$ \\
NTS-NOTEARS          & $0.287$ & $0.638$ & $29.5$ \\
VARLiNGAM            & $0.213$ & $0.703$ & $45.8$ \\
\cdnots{} + tetrad   & $0.277$ & $0.596$ & $32.5$ \\
\cdnots{}            & $0.198$ & $0.600$ & $41.2$ \\
DYNOTEARS            & $0.130$ & $0.721$ & $67.1$ \\
\bottomrule
\end{tabular}

\vspace{2pt}
{\footnotesize $^{\dagger}$LPCMCI is evaluated over the nine non-dense families
(Section~\ref{sec:lpcmci}).}
\end{table}

\subsection{LPCMCI and TS-ICD: latent-confounder-aware baselines}
\label{sec:lpcmci}

LPCMCI~\citep{gerhardus2020high} and TS-ICD~\citep{rohekar2023temporal} are the
baselines closest in intent to \lucid{}: both model latent confounders explicitly and
return a partial ancestral graph (PAG) over lagged \emph{and} contemporaneous edges.
Projecting a PAG onto the binary, lag-resolved edge set scored here discards information
both methods are designed to preserve, so neither $F_1$ should be read as a complete
assessment. We use the same projection rule for both, so they are scored identically: an
edge counts iff the mark at the effect is an arrowhead and the mark at the cause is not
(\texttt{-->} or \texttt{o->}; \texttt{<->} is excluded).

Both are computationally demanding here (Appendix~\ref{app:runtime}). For LPCMCI, even
with the possible-d-separating set capped at \texttt{max\_pds\_set=10}, orientation over
the near-complete PAG on pervasive-dense ($d=24$) exceeds our $10$\,GB memory budget; we
therefore report it on the remaining nine families, where it reaches
$F_1^{\text{dir}}=0.355$ (precision $0.233$, recall $0.671$, SHD $22.6$) against
PCMCI+'s $0.376$ on the same nine. TS-ICD has no analogous cap and is run on all ten,
reaching $0.133$ (precision $0.168$, recall $0.123$, SHD $14.0$), lower than every other
baseline in Table~\ref{tab:ood}. It is competitive only on the volatility- and
structural-confounding families (pervasive-vol $0.352$, GARCH $0.317$, pervasive-struct
$0.240$) and close to non-detection on the sparse topologies (Sparse-SF $0.007$), where
its output is mostly \texttt{<->}. We view both comparisons cautiously: they reflect the
operating regime and our binary lag-resolved evaluation, which is not the native output
space of PAG-based discovery.

\subsection{Router diagnostics}
\label{sec:diagnostics}

End-to-end performance does not reveal whether the router selects the intended regime. We
therefore evaluate it separately on a fixed diagnostic battery of $30$ sparse-regime and
$25$ pervasive-regime datasets ($d=15$), without retuning $\gamma$
(Appendix~\ref{app:router_sweeps}). At the default $\gamma=1.3$, the router detects all
$25$ pervasive cases (false-negative rate $0/25$) while routing $10/30$ sparse cases to
the pervasive branch.

Classification error need not translate directly into graph error because the two branches
can perform similarly on some datasets. On a separate $20$-seed rerun of the full OOD
suite, forcing all datasets through the sparse or pervasive branch gives family-weighted
$F_1$ of $0.337$ and $0.576$, respectively, compared with $0.592$ under automatic
routing. A per-dataset oracle that selects the better branch reaches $0.600$, leaving only
$0.007$ regret for the router. Branch choice nevertheless matters: the two fixed branches
differ by as much as $0.63$ on an individual family.

The router is therefore most useful as protection against regime misspecification. Relative
to the better fixed strategy (always pervasive), routing gains $0.017$ overall, with most
of the improvement on Sparse-ER ($0.655\to0.795$); Sparse-SF
($0.577\to0.593$) and Sparse-SW ($0.832\to0.843$) are nearly unchanged. More importantly,
routing avoids the much larger failure of committing to the wrong branch---$0.337$ for the
always-sparse strategy---without knowing the regime in advance.

\paragraph{Why \lucid{} does not filter lags $\ell\ge1$.}
We compare the default, which leaves pervasive-branch lagged edges unchanged, with
unconditional application of the volatility-based lag filters
(Appendix~\ref{app:filterablation}). On the full OOD suite ($20$ seeds), unconditional
filtering reduces family-weighted $F_1$ from $0.592$ to $0.451$, with particularly large
losses on homoskedastic families. We therefore restrict the pervasive correction to the
lag-$0$ slice.

\paragraph{Transient confounding and noise robustness.}
Two additional sweeps relax assumptions held fixed in the main benchmark. First, we vary a
latent factor's duty cycle from persistent to bursty using a two-state Markov switch
(Appendix~\ref{app:transient}). \lucid{} leads at four of five duty cycles and improves
from $0.623$ under full persistence to $0.818$ at duty cycle $0.05$, where its margin over
the fixed tetrad filter reaches $+0.22$. Second, we resimulate a confounded generator under
eight idiosyncratic noise distributions
(Appendix~\ref{app:noise}). \lucid{} leads across all tested distributions.
The lower-share setting is the more informative test of noise shape, since a dominant
Gaussian factor can mask the idiosyncratic distribution in the observed series.

\subsection{A market-calibrated semi-synthetic case study}
\label{sec:cds}

Purely synthetic benchmarks provide ground truth but risk depending on hand-specified
confounding dynamics; real observational data provide realistic dynamics but no known causal
graph. We bridge the two with a \emph{calibrate-then-simulate} design
(Appendix~\ref{app:calibrate}): we estimate the parameters of a factor-GARCH model from a
real market panel, simulate new series from the fitted model, inject known lag-$1$ edges,
and score their recovery. Thus the causal ground truth remains synthetic, while the
confounding dynamics are calibrated to real data.

We calibrate to the $11$-bank CDS panel taken from~\citet{fesanghary2025sypi},
using daily spreads over 2004--2008. The fitted model uses $\hat{k}=3$ latent factors and
GARCH(1,1)-$t$ dynamics, motivated by the strong heavy tails in the empirical residuals
(excess kurtosis $20$--$30$ under a Gaussian fit). We inject two lag-$1$ edges per
confounded group with weight $0.4$ and evaluate at $T=242$, matching the weekly-sampled
panel length ($10$ seeds), and at $T=1000$ for higher statistical power ($50$ seeds).

The \lucid{} layer leads at both sample sizes and improves both engines it wraps
(Table~\ref{tab:cds}). The strongest unwrapped baseline is PCMCI+ ($0.093$ at $T=1000$);
wrapping it yields $0.441$ (paired Wilcoxon $p=4\times10^{-9}$) at unchanged recall,
raising precision $0.050\!\to\!0.404$ and cutting SHD $25.1\!\to\!4.6$, while wrapping
\cdnots{} yields $0.080\!\to\!0.428$ ($p=2\times10^{-8}$). The two wrapped engines are
indistinguishable ($p=0.29$), consistent with the layer rather than the base engine
driving the gain. Seed variance is substantial, so we rely on the paired tests rather
than on any single mean gap.

\begin{table}[t]
\centering
\caption{CDS-calibrated semi-synthetic benchmark. Directed, lag-resolved
$F_1^{\text{dir}}$ (mean$\pm$std) measures recovery of injected lag-$1$ edges from a
$3$-factor GARCH(1,1)-$t$ model calibrated to a real $11$-bank CDS panel. $T=242$
matches the weekly-sampled panel length; $T=1000$ provides greater statistical power.
Precision (P), recall (R), and SHD are reported at $T=1000$. Each \lucid{} row is placed
beneath its corresponding base engine; \textbf{bold} marks the best entry per column.}
\label{tab:cds}
\resizebox{\columnwidth}{!}{%
\begin{tabular}{lccccc}
\toprule
 & $T=242$ & $T=1000$ & P & R & SHD \\
 & {\scriptsize(10 seeds)} & {\scriptsize(50 seeds)} & & & \\
\midrule
\cdnots{}            & $0.094${\scriptsize$\pm.048$} & $0.080${\scriptsize$\pm.052$} & $0.044$ & $0.530$ & $24.0$ \\
\lucid{} (\cdnots{}) & $0.502${\scriptsize$\pm.224$} & $0.428${\scriptsize$\pm.322$} & $0.442$ & $0.530$ & $4.4$ \\
\midrule
PCMCI+               & $0.099${\scriptsize$\pm.055$} & $0.093${\scriptsize$\pm.056$} & $0.050$ & $0.610$ & $25.1$ \\
\lucid{} (PCMCI+)    & $\mathbf{0.538}${\scriptsize$\pm.263$} & $\mathbf{0.441}${\scriptsize$\pm.303$} & $0.404$ & $0.610$ & $4.6$ \\
\bottomrule
\end{tabular}}
\end{table}

\section{Conclusion}
\label{sec:conclusion}

Latent confounding does not have a single statistical form, and correcting for the wrong
form can be as damaging as ignoring confounding altogether. \lucid{} addresses this with a
regime-adaptive deconfounding layer: a Mar\v{c}enko--Pastur router determines whether the
data support pervasive factor structure, and only then invokes spectral deconfounding and
lag-$0$ recovery; otherwise the sparse branch avoids an inappropriate low-rank correction.
Because this logic is separated from the causal search itself, the same layer can wrap
different discovery engines.

Across ten out-of-distribution confounding families, the default \lucid{} reaches
family-weighted directed, lag-resolved $F_1=0.596$, versus $0.404$ for the strongest
unwrapped baseline, and improves PCMCI+ and NTS-NOTEARS when wrapped around them as well.
The gains persist under transient and heavy-tailed confounding and in a market-calibrated
semi-synthetic study, with all defaults fixed. These results suggest adaptive
deconfounding is a useful layer around causal discovery rather than a replacement for it.

\paragraph{Limitations.}
\label{sec:limitations}
\lucid{} inherits the limits of spectral deconfounding. When genuine lag-$0$ structure
overlaps strongly with factor-dominated directions, spectral trimming can attenuate causal
signal together with confounding, as in the difficult \texttt{pervasive-dense} family. The
persistence gate reduces but cannot eliminate moralization, so S$-$L returns a skeleton
candidate rather than an identified skeleton, and its orientation rule is heuristic: the
lead-lag asymmetry weakens without sufficient temporal persistence.
The router relies on second-order spectral structure. Nonlinear factor mixing, or factors
too weak to produce a separated spectral spike, can therefore be misclassified
(Appendix~\ref{app:sensitivity}). Heavy-tailed idiosyncratic noise also
reduces accuracy: under Student-$t_3$ innovations, \lucid{} loses $0.050$ relative to the
Gaussian setting (Appendix~\ref{app:noise}). Finally, our evaluation remains synthetic or
semi-synthetic---even in the CDS study the scored series and edges are simulated---so
validation against real-world causal ground truth remains future work.

\section*{Impact Statement}

\lucid{} infers causal structure from observational time series, and like all
such methods it delivers a graph only under assumptions---here, that latent
confounding is either sparse and local or well approximated by a low-rank factor
structure, and that the relevant dynamics are captured at the analysis lag
horizon. When those assumptions do not hold, the output remains a graph and
carries no indication that it has ceased to be a causal one. Discovered edges
should therefore be treated as hypotheses to be corroborated with domain
knowledge, sensitivity analysis, or interventional evidence before being used to
support a causal claim, particularly in settings where such claims inform
consequential decisions.

\bibliographystyle{icml2025}
\bibliography{references}

\appendix
\onecolumn

\section{Additional Method and Experimental Details}
\label{app:method}

This appendix specifies the hyperparameters, benchmark families, implementation details,
and supporting ablations used in the main paper. All settings are fixed across the
evaluation suite unless explicitly stated otherwise.

\begin{table}[H]
\centering
\caption{Hyperparameters used by \lucid{}. $L$ is a standard lag-horizon input; all
remaining values are fixed across datasets and experiments. The persistence width
$\tau_g$ is the only value selected empirically, using generators excluded from the
ten-family OOD benchmark.}
\label{tab:params}
\footnotesize
\setlength{\tabcolsep}{4pt}
\begin{tabular}{p{4.0cm}p{6.6cm}lp{3.4cm}}
\toprule
Symbol & Role & Value & Source \\
\midrule
$L$ & analysis lag horizon & dataset-dep. & required input \\
\cdnots{} CI $\alpha$ & skeleton-discovery significance & $0.05$ & conventional \\
S$-$L level $\alpha$ & lag-$0$ edge-free-null level (\S\ref{sec:sl}) & $0.05$ & conventional \\
S$-$L draws $B$ & null draws for S$-$L calibration & $200$ & fixed \\
$\tau_g$ (Eq.~\ref{eq:slscore}) & persistence-gate width (\S\ref{sec:sl}) & $0.15$ & held-out selection \\
Orientation lag window & lag used by lead-lag asymmetry (\S\ref{sec:sl}) & $1$ & structural \\
$k$ & number of top eigenvalues in router statistic $R$ & $2$ & structural \\
$\gamma$ (Eq.~\ref{eq:mp}) & MP-router margin & $1.3$ & fixed \\
PDS $\alpha$ (sparse) & PDS significance level & $10^{-10}$ & fixed \\
HAC lag & Newey--West lag & $2$ & fixed \\
Max.\ selected controls & cap on Lasso-selected controls & $20$ & fixed \\
Factor-count margin & multiple of $\lambda_+$ (Eq.~\ref{eq:factorcount}) & $1.02$ & fixed \\
\bottomrule
\end{tabular}
\normalsize
\end{table}

\subsection{Out-of-distribution family definitions}
\label{app:families}

The OOD benchmark contains ten equally weighted confounding families. Three sparse
families use Erd\H{o}s--R\'enyi, scale-free, and small-world observed graphs with local
latent forks, spanning a range of sizes and confounding densities. Seven pervasive
families vary distinct properties of the latent process: \emph{pervasive-vol} varies
factor volatility through stochastic-volatility, regime-switching, static, and AR
factors; \emph{pervasive-struct} varies overlapping, signed, nonlinear, and heavy-tailed
loading structures; \emph{garch} uses the GARCH(1,1) reference process;
\emph{pervasive-dense} uses dense factor loadings at $d=24$, $T=2000$;
\emph{pervasive-mixedlag} varies mixed and delayed lag structure;
\emph{pervasive-heterog} varies factor-loading magnitudes; and
\emph{pervasive-nearunitroot} tests highly persistent latent dynamics.

Table~\ref{tab:dgp} summarizes each family's configuration. It is derived from the same
problem registry in the supplementary code that produced the reported results, which
additionally fixes the noise distributions and seeds. Note that $d$ is the \emph{observed}
dimension: on the sparse families the latent fork variables are excluded from the panel,
so $d$ is smaller than the nominal size in the configuration name.

\begin{table*}[b]
\centering
\caption{Data-generating configuration of the ten out-of-distribution families. Ranges
span the configurations within a family (``cfg'' counts them); $L$ is the ground-truth
maximum lag and $d$ the observed dimension. Loadings are drawn i.i.d.\ per entry.
Only \texttt{pervasive-dense} and \texttt{pervasive-mixedlag} contain ground-truth lag-$0$
edges ($158$ and $9$ respectively at seed $0$); the other eight families have none.}
\label{tab:dgp}
\resizebox{\textwidth}{!}{%
\begin{tabular}{lccccclll}
\toprule
Family & $d$ & $T$ & $L$ & cfg & $k$ & Loadings & Factor dynamics & Confounding strength \\
\midrule
Sparse-ER      & $9$--$21$  & $600$--$1500$  & $3$      & $4$ & local forks & --- & --- & confounded fraction $0.3$--$0.5$ \\
Sparse-SF      & $12$--$16$ & $1000$--$1500$ & $3$      & $2$ & local forks & --- & --- & confounded fraction $0.5$ \\
Sparse-SW      & $11$--$14$ & $600$--$800$   & $3$      & $2$ & local forks & --- & --- & confounded fraction $0.3$--$0.4$ \\
\midrule
Perv-vol       & $12$ & $1000$ & $1$      & $4$ & $3$ & $\mathcal{U}(0.5,1.5)$, block-disjoint & stoch.\ vol / regime-switch / static / AR & $w=0.4$, $2$ edges/group, $\rho_{\text{obs}}=0.3$ \\
Perv-struct    & $12$ & $1000$ & $1$      & $6$ & $3$ & as above; overlap / signed / nonlinear / heavy-tail & static & as above \\
GARCH          & $12$ & $1000$ & $1$      & $1$ & $3$ & $\mathcal{U}(0.5,1.5)$, block-disjoint & GARCH(1,1) & as above \\
Perv-dense     & $24$ & $2000$ & $1$      & $8$ & $4$ & $\pm\mathcal{U}(0.30,0.60)$, dense (all $d$) & AR(2) / GJR-GARCH / HMM break / nonlinear / drift & per-edge $0.16$--$0.35$ \\
Perv-mixedlag  & $15$ & $2000$ & $3$--$4$ & $4$ & $3$ & $\mathcal{U}(0.5,1.5)$, block-disjoint & static / GARCH / AR & $w=0.4$; lags $\{1,2,4\}$, delay $\le 3$ \\
Perv-heterog   & $15$ & $2000$ & $1$      & $3$ & $3$ & $\mathcal{U}(0.5,1.5)$; $30\%$ weak, $50\%$ moderate & static / GARCH / regime-switch & cancellation $\delta=0.05$ \\
Perv-nearunit  & $15$ & $2000$ & $1$      & $5$ & $3$ & $\mathcal{U}(0.5,1.5)$, block-disjoint & AR, $\rho_{\text{obs}},\rho_f\in\{0.3,0.95,0.99\}$ & $w=0.4$, $2$ edges/group \\
\bottomrule
\end{tabular}}
\end{table*}

\subsection{Pair-level evaluation}
\label{app:pairf1}

Prior time-series causal-discovery work often collapses predictions across lags and scores
an ordered variable pair as recovered whenever an edge is placed at any lag. We therefore
re-score the same predictions using this more permissive pair-level metric
$F_1^{\text{pair}}$. Self-loops remain excluded exactly as in the lag-resolved evaluation.

The qualitative conclusion is unchanged (Table~\ref{tab:ood_pair}), but collapsing lag
information reduces the separation between \lucid{} and the baselines. The default
\lucid{} changes from $0.596$ lag-resolved to $0.614$ pair-level, a gain of only $0.018$,
whereas the unwrapped baselines gain $0.043$--$0.082$. Against the strongest unwrapped
baseline, the margin is therefore $+0.16$ pair-level versus $+0.19$ when the predicted
lag must also be correct.

\begin{table*}[t]
\centering
\caption{OOD graph $F_1^{\text{pair}}$ under lag-collapsed scoring, using the same runs
as Table~\ref{tab:ood} ($50$ seeds; mean$\pm$std for Overall). An ordered pair is credited
when an edge is predicted at any lag. Bolding follows the same significance procedure as
Table~\ref{tab:ood}.}
\label{tab:ood_pair}
\resizebox{\textwidth}{!}{%
\begin{tabular}{lccccccccccc}
\toprule
Method & Overall & Sparse-ER & Sparse-SF & Sparse-SW & Perv-vol & Perv-struct & GARCH & Perv-dense & Perv-mixedlag & Perv-heterog & Perv-nearunit \\
\midrule
\cdnots{}                & $0.400${\scriptsize$\pm.017$} & $0.635$ & $0.527$ & $0.822$ & $0.370$ & $0.362$ & $0.376$ & $0.162$ & $0.277$ & $0.280$ & $0.186$ \\
\cdnots{} + tetrad       & $0.462${\scriptsize$\pm.020$} & $0.637$ & $0.578$ & $0.822$ & $0.458$ & $0.426$ & $0.428$ & $0.164$ & $0.366$ & $0.419$ & $\mathbf{0.316}$ \\
\lucid{} (\cdnots{}, default) & $\mathbf{0.619}${\scriptsize$\pm.025$} & $0.812$ & $0.623$ & $0.844$ & $0.626$ & $\mathbf{0.783}$ & $\mathbf{0.878}$ & $0.359$ & $\mathbf{0.382}$ & $\mathbf{0.544}$ & $0.344$ \\
\midrule
VARLiNGAM                & $0.402${\scriptsize$\pm.019$} & $0.641$ & $0.360$ & $0.775$ & $0.381$ & $0.372$ & $0.429$ & $0.222$ & $0.218$ & $0.426$ & $0.197$ \\
\midrule
LPCMCI$^{\dagger}$       & $0.437${\scriptsize$\pm.025$} & $0.571$ & $0.493$ & $0.747$ & $0.441$ & $0.394$ & $0.398$ & --- & $0.354$ & $0.333$ & $0.200$ \\
TS-ICD$^{\ddagger}$      & $0.200${\scriptsize$\pm.026$} & $0.058$ & $0.009$ & $0.066$ & $0.534$ & $0.401$ & $0.481$ & $0.006$ & $0.082$ & $0.189$ & $0.173$ \\
PCMCI+                   & $0.435${\scriptsize$\pm.016$} & $0.510$ & $0.526$ & $0.698$ & $0.432$ & $0.397$ & $0.436$ & $0.233$ & $0.333$ & $0.503$ & $0.282$ \\
\lucid{} (PCMCI+)        & $\mathbf{0.622}${\scriptsize$\pm.024$} & $0.822$ & $0.623$ & $\mathbf{0.869}$ & $\mathbf{0.649}$ & $0.767$ & $0.829$ & $\mathbf{0.401}$ & $0.360$ & $\mathbf{0.545}$ & $\mathbf{0.354}$ \\
\midrule
DYNOTEARS                & $0.292${\scriptsize$\pm.013$} & $0.309$ & $0.232$ & $0.298$ & $0.376$ & $0.363$ & $0.410$ & $0.209$ & $0.291$ & $0.252$ & $0.183$ \\
\midrule
NTS-NOTEARS              & $0.461${\scriptsize$\pm.016$} & $\mathbf{0.852}$ & $0.682$ & $\mathbf{0.861}$ & $0.341$ & $0.363$ & $0.413$ & $0.219$ & $0.298$ & $0.379$ & $0.199$ \\
\lucid{} (NTS-NOTEARS)   & $0.552${\scriptsize$\pm.022$} & $0.805$ & $\mathbf{0.712}$ & $0.832$ & $0.497$ & $0.579$ & $0.538$ & $0.320$ & $\mathbf{0.380}$ & $0.485$ & $\mathbf{0.371}$ \\
\bottomrule
\end{tabular}}

\vspace{2pt}
{\footnotesize $^{\dagger}$LPCMCI is not run on pervasive-dense ($d=24$); its Overall
score is family-weighted over the remaining nine families, where PCMCI+ scores $0.458$.
$^{\ddagger}$TS-ICD's PAG is projected with the same rule as LPCMCI's
(Section~\ref{sec:lpcmci}).}
\end{table*}

\subsection{Computational cost}
\label{app:runtime}

\lucid{} adds modest computational overhead relative to the wrapped discovery engine.
Across the three engine configurations in Table~\ref{tab:runtime}, the layer alone has
median runtime $0.6$--$0.7$\,s and maximum runtime $1.8$--$3.5$\,s, against base-engine
medians of $1.1$ to $31.4$\,s. On the dense stress case ($d=24$, $T=2000$) its median
cost is $1.2$\,s, compared with $17.4$\,s for \cdnots{}, $104.2$\,s for NTS-NOTEARS and
$143.2$\,s for PCMCI+. The $B=200$ edge-free-null draws account for most of this
overhead. The layer is implemented in NumPy and runs on a single CPU core, requiring no
accelerator even when the wrapped engine uses a GPU.

Its runtime distribution is also substantially narrower than those of the discovery
engines and the latent-aware PAG baselines: the layer stays below $3.5$\,s in every
reported run, whereas the slowest LPCMCI and TS-ICD instances take $4.1$ and $14.4$
hours. The additional cost of regime selection and deconfounding is therefore small
relative to causal search itself on the problems evaluated here.

\begin{table}[t]
\centering
\caption{Per-dataset runtime in seconds, pooled over the ten scored families and $50$
seeds. ``$+$ \lucid{} layer'' rows give the cost of the layer alone: each reuses its base
engine's stored skeleton, so base discovery is excluded. All runs used identical machines
(10-core CPU, one core per dataset, NVIDIA L4). \cdnots{} and NTS-NOTEARS are GPU-backed;
the remaining methods, and the \lucid{} layer itself, are CPU-only, so differences between
engines reflect implementation as well as algorithm.}
\label{tab:runtime}
\setlength{\tabcolsep}{4pt}
\begin{tabular}{lrrrr}
\toprule
Method & Median & Mean & p95 & Max \\
\midrule
\cdnots{}                & $1.1$ & $8.0$ & $59.6$ & $161$ \\
\quad $+$ \lucid{} layer & $0.7$ & $0.7$ & $1.3$ & $2.5$ \\
PCMCI+                   & $3.3$ & $61.6$ & $424.7$ & $1206$ \\
\quad $+$ \lucid{} layer & $0.7$ & $0.7$ & $1.4$ & $3.5$ \\
NTS-NOTEARS              & $31.4$ & $47.6$ & $140.3$ & $507$ \\
\quad $+$ \lucid{} layer & $0.6$ & $0.7$ & $1.3$ & $1.8$ \\
\midrule
LPCMCI$^{\dagger}$       & $8.0$ & $63.0$ & $48.2$ & $14{,}621$ \\
TS-ICD                   & $2.7$ & $223.9$ & $342.5$ & $51{,}875$ \\
DYNOTEARS                & $2.9$ & $13.6$ & $82.7$ & $201$ \\
VARLiNGAM                & $0.9$ & $1.4$ & $3.6$ & $4.3$ \\
\bottomrule
\end{tabular}

\vspace{2pt}
{\footnotesize $^{\dagger}$LPCMCI is not run on pervasive-dense, so its figures cover
$1{,}550$ datasets rather than $1{,}950$.}
\end{table}

\subsection{CDS calibrate-then-simulate procedure}
\label{app:calibrate}

For the CDS case study (Section~\ref{sec:cds}), we use the observed market panel to calibrate a simulation model. We VAR(1)-residualize the series, extract latent factors by PCA of the residual correlation matrix, and fit GARCH(1,1)-$t$ and AR(1)+noise models to the factor and idiosyncratic components, respectively. We then simulate new panels and inject known lag-$1$ causal edges, providing ground truth for evaluating causal recovery under market-calibrated dynamics.

\subsection{Spectral factor-count estimation}
\label{app:factorcount}

Let $C\in\RR^{d\times d}$ be the correlation matrix of the VAR(1) innovations, and let
$\lambda_1\ge\dots\ge\lambda_d$ denote its eigenvalues. For
$q=d/T_{\text{eff}}$, the Mar\v{c}enko--Pastur upper edge under the no-factor reference is

$$
\lambda_+=(1+\sqrt{q})^2.
$$

We estimate the number of pervasive spectral factors as
\begin{equation}
\hat{k}
=
\left|\left\{i:\lambda_i>1.02\,\lambda_+\right\}\right|,
\label{eq:factorcount}
\end{equation}
where the $1.02$ margin avoids counting eigenvalues immediately above the estimated bulk
edge. On the pervasive branch, the trim count is floored at one, so at least one direction
is suppressed whenever a dataset is routed to that branch.

This is a simple spectral factor-count estimator in the spirit of the factor-number
literature~\citep{bai2002determining,onatski2010determining,ahn2013eigenvalue}, specialized
to the MP edge. On the pervasive branch, $\hat{k}$ determines the number of leading
singular directions trimmed by S$-$L (Section~\ref{sec:sl}).

\subsection{Lag-$0$ orientation}
\label{app:orient}

The S$-$L stage recovers undirected lag-$0$ candidates, which must be oriented for directed
evaluation. Pairwise LiNGAM~\citep{hyvarinen2010estimation} provides one possible rule by
exploiting non-Gaussian asymmetry, but does not identify direction in the Gaussian case.
Because several benchmark families are Gaussian by construction, we instead use the
lead-lag statistic from Section~\ref{sec:sl},

$$
A(a,b)=
\frac{
|\operatorname{corr}(a_t,b_{t+1})|
-
|\operatorname{corr}(b_t,a_{t+1})|
}{
|\operatorname{corr}(a_t,b_{t+1})|
+
|\operatorname{corr}(b_t,a_{t+1})|
}.
$$

For a candidate pair, $A(a,b)>0$ gives $a\to b$ and $A(a,b)<0$ gives $b\to a$.

The motivation is specific to a persistent structural time-series model: an instantaneous
effect entering the target at time $t$ can propagate through the target's own dynamics,
creating a directional lag-$1$ asymmetry even under Gaussian innovations. This is a
heuristic, not a general orientation result. With little temporal persistence, the
propagated signal can vanish; more generally, predictive lead-lag asymmetry alone does not
establish causal direction outside the assumed SVAR setting.

The rule was selected on a separate battery of structural generators with contemporaneous
edges inserted inside the recursion. None of those datasets is part of the reported OOD
evaluation.

\subsection{PDS edge filter}
\label{app:pds}

For a candidate edge $(c,e,\ell)$ surviving skeleton discovery, at any lag
$\ell\in\{0,\dots,L\}$, let $Z\in\RR^{(T-L)\times dL}$ contain lags $1,\dots,L$ of all
observed variables except the candidate regressor, aligned with the target
$Y=x^e_{L+1:T}$. We run two Lasso regressions with BIC-selected regularization: one of $Y$
on $Z$, and one of the candidate $x^c_{t-\ell}$ on $Z$. The union of selected controls,
capped at $20$, defines $Z_{\text{sel}}$. We then fit
\begin{equation}
x^e_t
=
\beta_0+\beta_1 x^c_{t-\ell}
+\boldsymbol{\beta}^{\top}Z_{\text{sel},t}
+u_t
\label{eq:pds}
\end{equation}
by OLS with Newey--West standard errors (HAC lag $2$) and test
$H_0:\beta_1=0$. The edge is removed when $p\ge\alpha$, with
$\alpha=10^{-10}$ (Table~\ref{tab:params}). PDS is applied only on the sparse branch;
pervasive-branch edges at $\ell\ge1$ are returned unchanged. If the regression fails
numerically, for example because of collinear selected controls, the filter abstains and
retains the edge.

We use PDS as a conservative post-selection edge filter, not as a general identification
procedure for latent confounding. For $\ell\ge1$, \eqref{eq:pds} is a Granger-style test
over a high-dimensional observed information set; for $\ell=0$, the same double-selection
construction gives a contemporaneous partial-regression test with lagged controls. In both
cases, the controls are \emph{observed}. The post-double-selection framework of
\citet{belloni2014inference} relies on approximate sparsity among relevant observed
controls; our time-series use is therefore a filtering adaptation, not a claim that latent
forks are identified. In particular, a latent fork $H_t\to x^c_t$, $H_t\to x^e_t$ is not
blocked by conditioning on observed lags. The filter's narrower role is to remove candidate
edges that do not survive adjustment for a sparse set of observed predictors; behavior
under genuinely unblocked local forks remains an empirical question.

\subsection{Lag-$0$ recovery ablation}
\label{app:lag0}

This experiment isolates the value of recovering, rather than discarding, contemporaneous
structure. We inject genuine directed lag-$0$ edges into a pervasive-factor generator and
compare three policies: \emph{blanket drop}, which removes the entire lag-$0$ slice;
\emph{keep all}, which retains every unresolved contemporaneous edge from the base
skeleton; and S$-$L recovery (Section~\ref{sec:sl}). Injected edges are placed across factor
groups to avoid the more ambiguous case in which a direct edge and its shared factor act
on exactly the same group.

\begin{table}[H]
\centering
\caption{Lag-$0$ handling ablation on injected genuine contemporaneous edges
($20$ seeds, averaged across five factor processes). $F_1$ is directed and lag-resolved;
lag-$0$ recall is computed only over the injected contemporaneous edges.}
\label{tab:lag0}
\begin{tabular}{lccc}
\toprule
Contamination & Arm & $F_1$ & Lag-$0$ recall \\
\midrule
Low ($0.2$)  & Blanket drop & $0.658$ & $0.000$ \\
& Keep all     & $0.163$ & $0.732$ \\
& S$-$L        & $\mathbf{0.776}$ & $\mathbf{0.980}$ \\
\midrule
High ($0.5$) & Blanket drop & $0.458$ & $0.000$ \\
& Keep all     & $0.148$ & $0.563$ \\
& S$-$L        & $\mathbf{0.719}$ & $\mathbf{0.948}$ \\
\bottomrule
\end{tabular}
\end{table}

Table~\ref{tab:lag0} shows the tradeoff directly. Blanket removal yields zero lag-$0$
recall. Keeping the unresolved slice recovers many genuine edges but also retains the
factor footprint, giving the lowest overall $F_1$. S$-$L attains the highest lag-$0$ recall
($0.980$ and $0.948$) and the highest overall $F_1$ ($0.776$ and $0.719$) at the two
contamination levels.

S$-$L can recover more genuine lag-$0$ edges than the keep-all arm because the candidate
sets differ: keep-all is limited to edges proposed by the original skeleton, whereas
S$-$L constructs candidates from the deconfounded precision matrix.

The generators used here, together with an overlapping-loading variant, were also used to
select the persistence width $\tau_g$. Table~\ref{tab:lag0} should therefore be interpreted
as a mechanism/development ablation rather than held-out evidence for that hyperparameter.
The ten-family OOD benchmark and the sensitivity analysis in
Appendix~\ref{app:sensitivity} provide the corresponding out-of-sample evaluation.

\subsection{Lag-$\ell\ge1$ filter ablation}
\label{app:filterablation}

We test whether pervasive-branch lagged edges benefit from additional volatility-specific
filtering. Table~\ref{tab:ablation} compares \lucid{} as shipped, which retains
pervasive-branch lagged edges from the base engine, with a variant that applies the
volatility-based filters unconditionally. On the full ten-family suite, filtering reduces
family-weighted $F_1$ from $0.592$ to $0.451$; restricted to the seven pervasive families,
the corresponding averages are $0.528$ and $0.323$. We therefore leave lags $\ell\ge1$
unchanged on the pervasive branch.

\begin{table}[H]
\centering
\caption{Ablation of pervasive-branch lag-$\ell\ge1$ filtering on the full ten-family OOD
suite ($20$ seeds). Overall is the ten-family weighted mean; the seven pervasive-family
columns are shown individually.}
\label{tab:ablation}
\footnotesize
\setlength{\tabcolsep}{4pt}
\begin{tabular}{lcccccccc}
\toprule
& Overall & P-vol & P-struct & GARCH & P-dense & P-mixedlag & P-heterog & P-nearunit \\
\midrule
\lucid{} (as shipped)           & $\mathbf{0.592}$ & $\mathbf{0.600}$ & $\mathbf{0.761}$ & $\mathbf{0.877}$ & $\mathbf{0.342}$ & $\mathbf{0.317}$ & $\mathbf{0.456}$ & $\mathbf{0.341}$ \\
$+$ unconditional filtering     & $0.451$ & $0.244$ & $0.387$ & $0.575$ & $0.150$ & $0.249$ & $0.424$ & $0.233$ \\
\bottomrule
\end{tabular}
\end{table}

\subsection{Robustness to transient confounding}
\label{app:transient}

To test intermittent confounding, we modulate a latent factor with a two-state Markov
process and vary its active duty cycle from persistent ($1.0$) to highly transient
($0.05$). Each dataset has $d=15$ and $T=1500$, with $20$ seeds per duty cycle.

\lucid{} has the highest $F_1$ at four of the five duty cycles
(Table~\ref{tab:transient}) and reaches its highest score at the most intermittent setting,
increasing from $0.623$ at duty $1.0$ to $0.818$ at duty $0.05$. At duty $0.25$, the routed
tetrad configuration performs best ($0.715$ versus $0.599$ for \lucid{}). The comparison
therefore does not show uniform dominance, but neither does it show systematic degradation
as confounding becomes more intermittent.

\begin{table}[H]
\centering
\caption{Directed, lag-resolved $F_1$ under transient factor confounding
($d=15$, $T=1500$, $20$ seeds) as a function of factor duty cycle. ``Routed (tetrad
base)'' uses the routed pipeline with tetrad-based pervasive correction and no S$-$L
lag-$0$ recovery. Best per column is bold.}
\label{tab:transient}
\begin{tabular}{lccccc}
\toprule
duty cycle & 1.00 & 0.50 & 0.25 & 0.10 & 0.05 \\
\midrule
\cdnots{}              & $0.161$ & $0.160$ & $0.130$ & $0.149$ & $0.177$ \\
Tetrad filter          & $0.487$ & $0.586$ & $0.660$ & $0.615$ & $0.596$ \\
Routed (tetrad base)   & $0.496$ & $0.649$ & $\mathbf{0.715}$ & $0.697$ & $0.701$ \\
\lucid{}               & $\mathbf{0.623}$ & $\mathbf{0.683}$ & $0.599$ & $\mathbf{0.705}$ & $\mathbf{0.818}$ \\
\bottomrule
\end{tabular}
\end{table}

\subsection{Robustness to the noise distribution}
\label{app:noise}

The main OOD generators use Gaussian idiosyncratic innovations. To isolate sensitivity to
the noise distribution, we hold one pervasive-factor generator fixed
($d=12$, $T=1000$, homoskedastic factor, injected lag-$1$ edges) and resimulate it under
eight idiosyncratic noise models: Gaussian, Laplace, Student-$t$ with $\nu=3$, uniform,
exponential, heteroscedastic location-scale, ARCH-type history-dependent, and
post-nonlinear noise.

Because the observed innovation combines common-factor and idiosyncratic components, a
dominant Gaussian factor can mask non-Gaussian idiosyncratic behavior. We therefore set
the factor communality to $c=0.3$, so the noise distribution remains visible in the
observed data. Table~\ref{tab:noise} reports directed, lag-resolved $F_1$ over $50$ seeds.

\begin{table}[H]
\centering
\caption{Robustness to idiosyncratic noise distribution on a fixed pervasive-factor
generator ($d=12$, $T=1000$, $50$ seeds, communality $c=0.3$). $t_3$ denotes
Student-$t$ noise with $\nu=3$; Het.\ is heteroscedastic location-scale, Hist.\ is
ARCH-type history dependence, and PNL is post-nonlinear noise. Best per column is bold.}
\label{tab:noise}
\footnotesize
\setlength{\tabcolsep}{4pt}
\begin{tabular}{lcccccccc}
\toprule
Method & Gaussian & Laplace & $t_3$ & Uniform & Exp. & Het. & Hist. & PNL \\
\midrule
\cdnots{}              & $0.269$ & $0.272$ & $0.266$ & $0.269$ & $0.270$ & $0.264$ & $0.268$ & $0.288$ \\
\cdnots{} + tetrad     & $0.280$ & $0.280$ & $0.279$ & $0.282$ & $0.280$ & $0.285$ & $0.286$ & $0.291$ \\
DYNOTEARS              & $0.282$ & $0.283$ & $0.282$ & $0.285$ & $0.285$ & $0.296$ & $0.294$ & $0.179$ \\
PCMCI+                 & $0.336$ & $0.337$ & $0.331$ & $0.336$ & $0.332$ & $0.344$ & $0.350$ & $0.342$ \\
LPCMCI                 & $0.250$ & $0.247$ & $0.271$ & $0.270$ & $0.282$ & $0.290$ & $0.298$ & $0.354$ \\
VARLiNGAM              & $0.380$ & $0.362$ & $0.360$ & $0.367$ & $0.366$ & $0.353$ & $0.333$ & $0.335$ \\
NTS-NOTEARS            & $0.287$ & $0.286$ & $0.283$ & $0.285$ & $0.284$ & $0.275$ & $0.277$ & $0.272$ \\
\lucid{} (PCMCI+)      & $0.481$ & $0.476$ & $0.430$ & $0.473$ & $0.469$ & $0.481$ & $0.488$ & $0.485$ \\
\lucid{}               & $\mathbf{0.489}$ & $\mathbf{0.485}$ & $\mathbf{0.438}$ & $\mathbf{0.488}$ & $\mathbf{0.479}$ & $\mathbf{0.511}$ & $\mathbf{0.515}$ & $\mathbf{0.487}$ \\
\bottomrule
\end{tabular}
\end{table}

\lucid{} has the highest $F_1$ in all eight noise conditions, with margins of
$0.078$--$0.144$ over the strongest non-\lucid{} baseline in each column. Heavy-tailed
Student-$t_3$ noise produces the largest degradation relative to the Gaussian condition:
\lucid{} decreases from $0.489$ to $0.438$, consistent with the limitation discussed in
Section~\ref{sec:limitations}.

\section{Reproducibility}
\label{app:repro}


All methods described in this paper are implemented in the open-source
\texttt{causal-ts} library~\citep{fesanghary2026causalts}, available at
\url{https://github.com/bloomberg/causal-ts}.

\subsection{Router diagnostics and sensitivity}
\label{app:router_sweeps}

The router diagnostic of Section~\ref{sec:diagnostics} contains $30$ sparse-regime
datasets (three graph topologies, with and without local latent forks; $5$ seeds each) and
$25$ pervasive-regime datasets (five factor processes, $5$ seeds each), at $d=15$. The
$15$ unconfounded and all pervasive datasets use $T=2000$, while the $15$
sparse-confounded datasets use $T=1000$. Sparse datasets are negatives and pervasive
datasets positives, so FPR denotes routing a sparse dataset to the pervasive branch and
FNR the converse. No value of $\gamma$ is retuned on this battery.

At the default $\gamma=1.3$, the router detects all pervasive cases (FNR $=0$) and
misroutes $10/30$ sparse cases (FPR $=0.333$). In the separate $20$-seed OOD routing
ablation, moving from always-pervasive to automatic routing changes Sparse-ER from $0.655$
to $0.795$, Sparse-SF from $0.577$ to $0.593$, and Sparse-SW from $0.832$ to $0.843$,
while the seven pervasive families are essentially unchanged. This is consistent with the
asymmetric cost reported in Section~\ref{sec:diagnostics}: missing pervasive confounding
can be more damaging than applying the pervasive branch to some sparse datasets. The
default margin therefore favors sensitivity to pervasive factors over classification
accuracy on this diagnostic battery alone.

Table~\ref{tab:router_sweeps} examines the router margin and its dependence on dimension
and sample size. Within this battery and the tested grid, larger $\gamma$ improves
balanced accuracy: FPR reaches $0$ by $\gamma=2.2$ while FNR remains $0$. Thus,
$\gamma=1.3$ is not the balanced-accuracy-maximizing choice on this diagnostic. It is a
fixed margin over the MP edge that was set independently of this battery; reselecting it
at $2.2$ from these $55$ datasets would instead tune to the diagnostic. Retaining the
smaller default also preserves sensitivity to weaker factors: in the factor-strength
sweep, FNR remains zero through loading range $(0.3,0.6)$ and rises as factors weaken
further.

The bottom panel uses only the $15$ unconfounded sparse datasets as negatives (and
\texttt{garch}/\texttt{static} as positives), so its $(15,2000)$ entry of $0.33$ is
computed on a different, smaller negative set than the $0.33$ default-$\gamma$ FPR of the
full $30$-dataset battery above. The two coincide numerically here but are not directly
comparable.

This is the expected difficulty for eigenvalue-based detection as a factor spike approaches
the noise bulk, consistent with spectral phase-transition behavior~\citep{baik2005phase}.
The main OOD benchmark uses the stronger loading range $(0.5,1.5)$; weak factors outside
this regime are therefore a documented limitation rather than part of the headline
benchmark.

\begin{table}[H]
\centering
\caption{Router diagnostics. \textbf{Top:} FPR and balanced accuracy versus the MP
margin $\gamma$; the default $\gamma=1.3$ is marked by $^\ast$.
\textbf{Bottom:} FPR across dimension and sample-size settings at $\gamma=1.3$.
FNR is zero for all configurations shown.}
\label{tab:router_sweeps}
\footnotesize
\setlength{\tabcolsep}{4pt}
\begin{tabular}{lccccccc}
\toprule
$\gamma$ & $0.8$ & $1.0$ & $1.15$ & $1.3^\ast$ & $1.5$ & $1.8$ & $2.2$ \\
\midrule
FPR & $1.00$ & $0.77$ & $0.43$ & $0.33$ & $0.17$ & $0.03$ & $0.00$ \\
Balanced acc. & $0.50$ & $0.62$ & $0.78$ & $0.83$ & $0.92$ & $0.98$ & $1.00$ \\
\bottomrule
\end{tabular}
\vspace{2pt}
\begin{tabular}{lcccccc}
\toprule
$(d,T)$ & $(10,500)$ & $(10,2000)$ & $(15,2000)$ & $(25,1000)$ & $(25,3000)$ & $(30,1000)$ \\
\midrule
FPR & $0.07$ & $0.13$ & $0.33$ & $0.20$ & $0.27$ & $0.27$ \\
\bottomrule
\end{tabular}
\end{table}

\paragraph{Number of top eigenvalues $k$.}
The statistic and its threshold both sum over the top $k$ eigenvalues, so under the
no-factor null each additional eigenvalue contributes approximately $\lambda_+$ to both,
and the ratio $R/\tau$ is nearly invariant to $k$. The asymptotic separation result
likewise requires only that $k$ be fixed, not that it equal $2$.
Table~\ref{tab:router_k} sweeps $k$ over the same battery. At the benchmark loading range,
FNR is $0$ for every $k$, while FPR falls monotonically from $0.40$ at $k=1$ to $0.13$ at
$k=4$. Thus, the shipped $k=2$ is not the balanced-accuracy-maximizing choice on this
battery; both $k=3$ and $k=4$ attain lower FPR at the same FNR. We retain $k=2$ because it
was fixed before this diagnostic existed rather than selecting it on the diagnostic. A
tradeoff appears at weaker factor strength: at loading range $(0.2,0.4)$, $k=4$ begins to
miss pervasive structure while $k\le3$ still detects all of it.

\begin{table}[H]
\centering
\caption{Router sensitivity to the number of top eigenvalues $k$, on the same
$55$-dataset battery as Table~\ref{tab:router_sweeps} ($\gamma=1.3$; every dataset is
scored at every $k$, so the comparison is paired). The $k=2^\ast$ column reproduces the
$\gamma=1.3$ entries of Table~\ref{tab:router_sweeps}. The last row reports FNR at the
weak loading range $(0.2,0.4)$; all other rows use the benchmark range $(0.5,1.5)$.}
\label{tab:router_k}
\footnotesize
\setlength{\tabcolsep}{4pt}
\begin{tabular}{lcccc}
\toprule
$k$ & $1$ & $2^\ast$ & $3$ & $4$ \\
\midrule
FPR                        & $0.40$  & $0.33$  & $0.20$  & $0.13$ \\
FNR                        & $0.00$  & $0.00$  & $0.00$  & $0.00$ \\
Balanced acc.              & $0.800$ & $0.833$ & $0.900$ & $0.933$ \\
FNR, loadings $(0.2,0.4)$  & $0.00$  & $0.00$  & $0.00$  & $0.16$ \\
\bottomrule
\end{tabular}
\end{table}

\subsection{Sensitivity to fixed constants}
\label{app:sensitivity}

We next vary the three downstream constants most likely to affect the reported results:
the sparse-branch PDS level, the S$-$L edge-free-null level, and the persistence-gate width
$\tau_g$. Because all three act after skeleton discovery, each sweep reuses the same cached
base skeleton and reruns the corresponding deconfounding step. The router margin $\gamma$
is studied separately above.

Table~\ref{tab:const_sensitivity} shows limited sensitivity to all three choices. The PDS
sweep ranges from $10^{-12}$ to a conventional $0.05$.
Family-weighted $F_1^{\text{dir}}$ remains stable from $10^{-12}$ through $10^{-3}$
($0.588$--$0.598$) and declines modestly at conventional testing levels, to $0.585$ at
$10^{-2}$ and $0.573$ at $0.05$, as precision falls from $0.659$ to $0.602$ and recall
rises from $0.606$ to $0.641$. The stringent default is therefore best viewed as a
conservative edge-inclusion criterion rather than a conventional hypothesis-testing level
(Appendix~\ref{app:pds}). At $\alpha=0.05$, the layer still reaches $0.573$, above the
$0.411$ of the strongest unwrapped baseline, so the overall performance advantage persists
without the stringent default. Varying the S$-$L null level from $0.01$ to $0.2$ changes
$F_1$ by only $0.002$. Neither default was selected to maximize OOD performance: PDS
$\alpha=10^{-4}$ scores marginally higher ($0.598$) than the shipped $10^{-10}$ ($0.592$).

The persistence width is similarly insensitive: sweeping $\tau_g$ from $0$ (gate off) to
$0.40$ changes family-weighted $F_1$ from $0.596$ to $0.589$, with $0.592$ at the default
$\tau_g=0.15$. Thus, the OOD result is not materially sensitive to the empirically selected
gate width. This sweep is a robustness check rather than evidence for the gate's benefit:
collider-induced moralization is uncommon in this broad OOD suite, so the errors targeted
by the gate are rare there.

We therefore evaluate the gate on a collider generator where its target error is common,
using disjoint triples $(X,Y,Z)$ and $Z_t=w(X_t+Y_t)+\varepsilon_t$, so $X$ and $Y$ are
co-parents with no direct edge ($d{=}12$, $T{=}120$). Scoring the recovered lag-$0$
skeleton by thresholded $F_1$---so surviving co-parents count as false positives---the
shipped gate gives $F_1=1.000$ against $0.957$ with the gate disabled (paired $+0.043$;
$13$ wins, $0$ losses, $7$ ties; Wilcoxon $p=6.3\times10^{-4}$ over $20$ seeds), and admits
$0.00$ co-parent false edges per dataset against $0.75$. This is a mechanism check rather
than held-out validation: the generator was one cell of the battery used to select
$\tau_g$, and only the seeds here are disjoint from selection. On this generator, the gate
removes the errors it targets; its near-zero effect on the OOD suite is consistent with
those errors being rare there.

\begin{table}[H]
\centering
\caption{Sensitivity of \lucid{} to downstream fixed constants
(family-weighted directed, lag-resolved $F_1$; $20$ seeds). One constant is varied at a
time with all others fixed at their defaults ($^\ast$). $\tau_g=0$ disables the
persistence gate.}
\label{tab:const_sensitivity}
\begin{tabular}{lccc}
\toprule
Value & $F_1$ & Precision & Recall \\
\midrule
\multicolumn{4}{c}{\emph{Sparse-branch PDS $\alpha$}} \\
$10^{-12}$ & $0.588$ & $0.659$ & $0.606$ \\
$10^{-10,\ast}$ & $0.592$ & $0.657$ & $0.613$ \\
$10^{-8}$  & $0.596$ & $0.655$ & $0.621$ \\
$10^{-6}$  & $0.597$ & $0.652$ & $0.627$ \\
$10^{-4}$  & $0.598$ & $0.646$ & $0.634$ \\
$10^{-3}$  & $0.596$ & $0.639$ & $0.638$ \\
$10^{-2}$  & $0.585$ & $0.618$ & $0.641$ \\
$0.05$     & $0.573$ & $0.602$ & $0.641$ \\
\midrule
\multicolumn{4}{c}{\emph{S$-$L edge-free-null $\alpha$}} \\
$0.01$  & $0.594$ & $0.661$ & $0.612$ \\
$0.025$ & $0.593$ & $0.659$ & $0.613$ \\
$0.05^{,\ast}$ & $0.592$ & $0.657$ & $0.613$ \\
$0.1$   & $0.592$ & $0.655$ & $0.614$ \\
$0.2$   & $0.592$ & $0.653$ & $0.616$ \\
\midrule
\multicolumn{4}{c}{\emph{Persistence-gate width $\tau_g$}} \\
$0$ (no gate) & $0.596$ & $0.659$ & $0.617$ \\
$0.05$  & $0.596$ & $0.659$ & $0.616$ \\
$0.10$  & $0.594$ & $0.658$ & $0.615$ \\
$0.15^{,\ast}$ & $0.592$ & $0.657$ & $0.613$ \\
$0.25$  & $0.590$ & $0.655$ & $0.613$ \\
$0.40$  & $0.589$ & $0.652$ & $0.612$ \\
\bottomrule
\end{tabular}
\end{table}

\end{document}